\documentclass[10pt, a4paper]{article}
\usepackage[utf8]{inputenc}
\usepackage{amsmath}
\usepackage{amsfonts}
\usepackage{amssymb}
\usepackage{graphicx}
\usepackage{hyperref}
\usepackage[a4paper, total={6in, 8in}]{geometry}
\usepackage{caption}
\usepackage{subcaption}
\usepackage{float}

\title{RL as Regressor: A Reinforcement Learning Approach for Function Approximation}
\author{Yongchao Huang \footnote{Author email: yongchao.huang@abdn.ac.uk}}
\date{\today}

\begin{document}

\maketitle

\begin{abstract}
Standard regression techniques, while powerful, are often constrained by predefined, differentiable loss functions such as mean squared error. These functions may not fully capture the desired behavior of a system, especially when dealing with asymmetric costs or complex, non-differentiable objectives. In this paper, we explore an alternative paradigm: framing regression as a Reinforcement Learning (RL) problem. We demonstrate this by treating a model's prediction as an action and defining a custom reward signal based on the prediction error, and we can leverage powerful RL algorithms to perform function approximation. Through a progressive case study of learning a noisy sine wave, we illustrate the development of an Actor-Critic agent, iteratively enhancing it with Prioritized Experience Replay, increased network capacity, and positional encoding to enable a capable RL agent for this regression task. Our results show that the RL framework not only successfully solves the regression problem but also offers enhanced flexibility in defining objectives and guiding the learning process.
\end{abstract}

\section{Introduction}
Function approximation is a cornerstone of machine learning, with regression models being the primary tool for predicting continuous numerical outcomes. The standard approach involves minimizing a loss function, typically mean squared error (MSE) or mean absolute error (MAE), between the model's predictions and ground truth values. This is effective for a wide range of problems, but it has limitations - one of which being, the choice of loss function is often a matter of convenience rather than a true reflection of the underlying objective. For example, in financial forecasting, overestimating a stock's price might be far more costly than underestimating it - a nuance that MSE treats symmetrically.

Reinforcement Learning (RL) offers a more flexible and powerful framework. At its core, RL is about training an agent to make a sequence of decisions to maximize a cumulative reward signal. What if we frame a single prediction as a decision? This reframing turns a regression task into a contextual bandit problem \footnote{Contextual bandits are a class of one-step reinforcement learning algorithms where the agent uses side information (context) to inform its action selection. They are widely used in applications like personalized recommendations and clinical trials \cite{langford_epoch-greedy_2007}.} - a special case of RL. The key advantages of this RL regressor approach are:
\begin{itemize}
    \item \textbf{Customizable, task-dependent objective:} the reward function can be arbitrarily complex and tailored to the specific goals of the task, even if it is non-differentiable. We can design rewards that heavily penalize certain types of errors while tolerating others.
    \item \textbf{Principled exploration:} RL agents inherently balance exploration (trying new actions to discover better strategies) and exploitation (using the current best strategy). This can help models escape local minima that might trap traditional gradient-based optimizers.
    \item \textbf{Advanced learning strategies:} the RL framework provides access to sophisticated techniques such as Prioritized Experience Replay (PER), which focuses the agent's attention on the most informative examples.
\end{itemize}

In this paper, we provide a tutorial-like walk-through of this RL for regression concept. We begin by outlining the methodology for transforming a regression problem into an RL task using an Actor-Critic architecture. We then present a progressive case study where we train an agent to learn a noisy sine function, starting with a simple version and iteratively improve the RL agent capacity. Through this example, we demonstrate how to diagnose model failings and systematically improve performance by incorporating advanced RL techniques and feature engineering, attaining in a accurate and robust RL model for this regression task.

\section{Related Work}
The idea of connecting supervised learning and reinforcement learning is not new. The work of Williams \cite{williams_simple_1992} on the REINFORCE algorithm laid the groundwork for using RL to train networks where the objective function is not directly differentiable with respect to the network's outputs. More recently, the success of deep RL in complex domains such as game playing \cite{mnih_human-level_2015} and robotics \cite{tang_deep_2024} has increased interest in applying RL techniques more broadly.

The formulation of regression as a contextual bandit problem is a direct application of this line of thinking. In contextual bandits, an agent chooses an action based on a given state (or context) to maximize an immediate reward. This perfectly maps to a regression setting where the state is the input features, the action is the prediction, and the reward is a function of the prediction's accuracy.

Our work draws heavily on Actor-Critic methods, particularly deep deterministic policy gradients (DDPG) \cite{lillicrap_continuous_2019}, which are well-suited for continuous action spaces as those found in regression. We also incorporate Prioritized Experience Replay \cite{schaul_prioritized_2016}, a technique that improves sample efficiency by replaying important transitions more frequently. Further, our use of positional encoding is inspired by its success in the \textit{Transformer} \cite{vaswani_attention_2017} architecture, where it was used to inject information about the order of sequence elements.

\section{Approach}
To transform a standard regression task into an RL problem, we must define the core components of an RL environment: the state, action, and reward.

\begin{itemize}
    \item \textbf{State ($S_t$):} the input features $X$ of a data point serve as the state. The state provides the context for the agent's decision.
    \item \textbf{Action ($A_t$):} the agent's numerical prediction, $\hat{y}$, is the action. Since this is a continuous value, we are operating in a continuous action space.
    \item \textbf{Reward ($R_t$):} the reward is a scalar value calculated from the error between the prediction $\hat{y}$ and the true outcome $y$. Instead of minimizing a loss, the agent's goal is to maximize this reward. We use the Gaussian kernel as our reward function:
    \begin{equation}
        R(\hat{y}, y) = \exp\left(-\frac{(y - \hat{y})^2}{2\sigma^2}\right)
    \end{equation}
    This function provides a smooth reward that is maximal (1.0) when the error is zero and decays gracefully as the error increases. The hyperparameter $\sigma$ controls the tolerance for error.
\end{itemize}

We implement this framework using an Actor-Critic architecture \footnote{Readers are free to use other RL algorithms such as Soft Actor-Critic (SAC), Twin Delayed DDPG (TD3), or Proximal Policy Optimization (PPO), which are all well-suited for tasks with continuous action spaces.}, which consists of two neural networks:
\begin{enumerate}
    \item \textbf{The Actor ($\pi(S)$):} this network represents the policy. It takes the state $S$ as input and outputs a deterministic action $A$ (the prediction $\hat{y}$).
    \item \textbf{The Critic ($Q(S, A)$):} this network represents the value function. It learns to predict the expected reward (the Q-value) for a given state-action pair.
\end{enumerate}

The training proceeds by alternating between updating the Critic and the Actor. The Critic is trained to minimize the difference between its predicted Q-value and the actual reward received. The Actor is then updated to produce actions that the Critic predicts will yield a higher reward.

\section{A Progressive Example}

We demonstrate our approach on a synthetic task: learning the function $y = \sin(x)$ from noisy samples. We progressively improve the learning process, and illustrate how to diagnose and overcome common failure modes. 

The Actor and Critic are both represented by 3 / 4-layer, fully-connected feedforward networks (MLPs) with ReLU activations; the Actor's output layer uses a \textit{Tanh} activation to bound the prediction, while the Critic's output layer is \textit{linear} to predict the unbounded Q-value.
Across all experiments, we use the \textit{Adam} optimizer \cite{kingma_adam_2017} for both networks. The learning rate is set to $1 \times 10^{-4}$ for the Actor and $1 \times 10^{-3}$ for the Critic. The models are trained on a dataset of 1000 samples using a batch size of 64. For exploration, we add Gaussian noise with a standard deviation of 0.1 to the Actor's actions during training. The reward function's sensitivity parameter (length-scale) $\sigma$ is set to 0.2. 
The total number of training epochs is 500 for all experiments.
Critic loss is calculated as the mean squared error between the value it predicts for a given state-action pair and the actual reward received, while the Actor loss is the negative of the Critic's value prediction, which is minimized to guide the Actor towards outputting actions that the Critic deems more rewarding.
Each experiment takes around 30s to run in Google Colab with CPU use only.

\subsection{A limited working example}
\begin{figure}[H]
    \centering
    \includegraphics[width=0.8\textwidth]{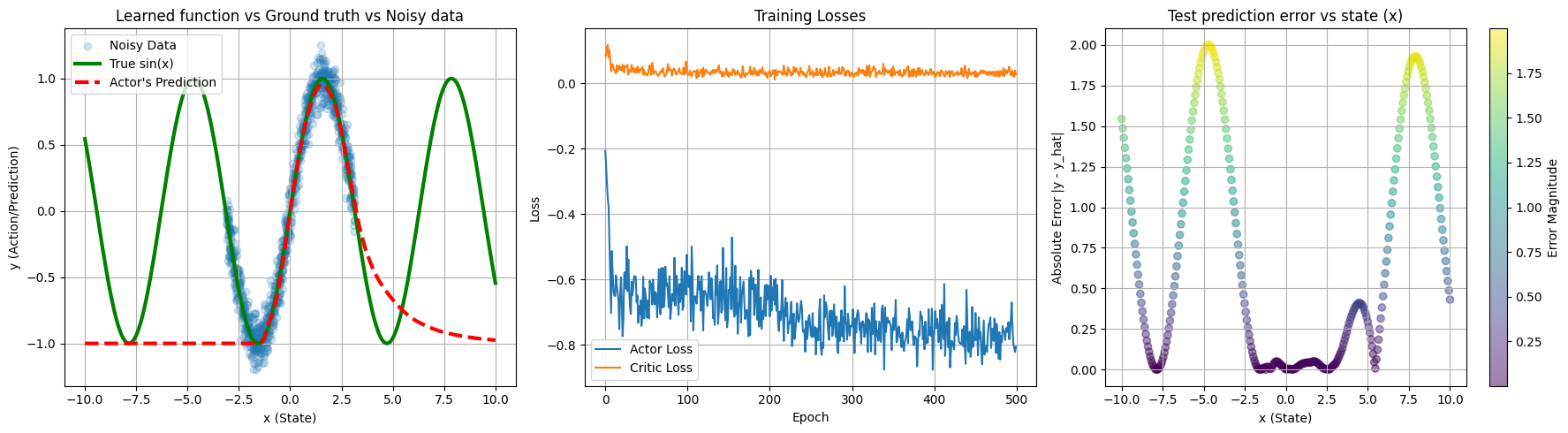}
    \caption{Stage 1 results. The model learns the function over a single period but fails to generalize.}
    \label{fig:stage1}
\end{figure}

First, we tackle a simple version of the problem, learning the function over a single period ($x \in [-\pi, \pi]$). 
We use a standard Actor-Critic setup where both the Actor and Critic are represented by 3-layer multi-layer perceptron (MLP) with two hidden layers containing 128 and 64 neurons respectively, using ReLU activations and a \textit{Tanh} activation on the Actor's final output to bound the prediction.
The results (Fig.\ref{fig:stage1}) show that the model learns partial of the function within training regime. However, it fails to generalize outside this narrow range, simply outputting a constant value. This initial success validates the core concept but highlights the limitations of the simple approach.

\subsection{More training data with prioritized experience replay}
\begin{figure}[H]
    \centering
    \includegraphics[width=0.8\textwidth]{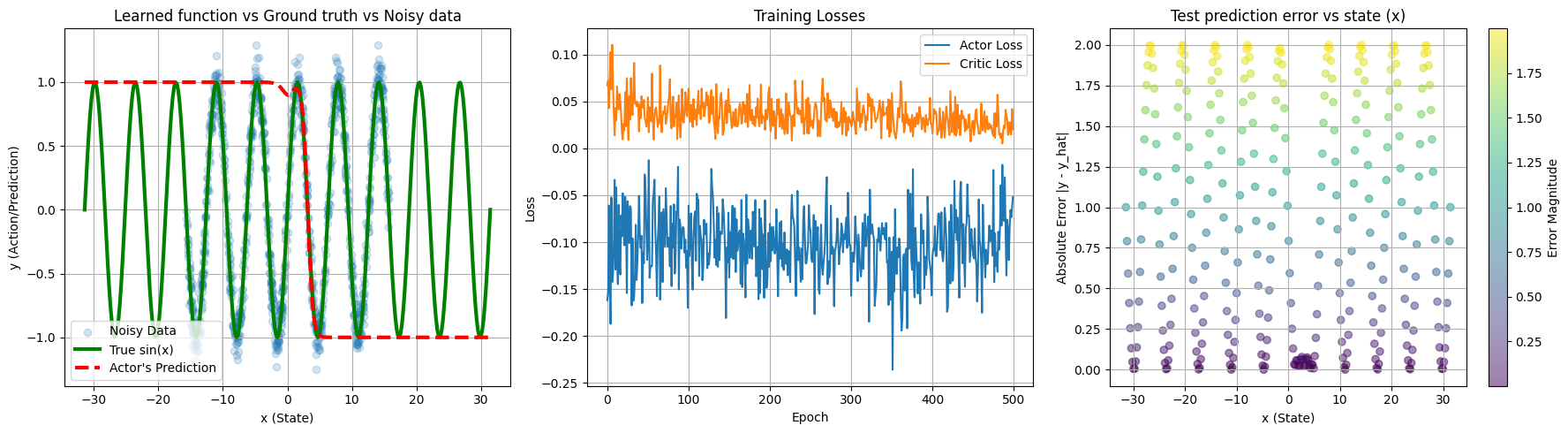}
    \caption{Stage 2 results. With an expanded data range, the simple model fails to learn the periodic function, indicating insufficient network capacity.}
    \label{fig:stage2}
\end{figure}

Next, we expand the training data range to five periods ($x \in [-5\pi, 5\pi]$), and keep the same model architecture and introduce Prioritized Experience Replay (PER) to help the agent focus on high-error regions. The results (Fig.\ref{fig:stage2}) are poor, and the training fails (Actor training losses even increase at the end). The model completely fails to capture the periodic nature of the data, collapsing to a simple step function. The error-state plot on the right confirms this, showing very high errors at the peaks and troughs of the sine wave. This demonstrates a classic case of underfitting: the model's network architecture lacks the capacity to represent the more complex target function.

\subsection{Increasing network capacity}
To address the underfitting, we increase the capacity of both the Actor and Critic networks, making them deeper and wider by adding a third hidden layer and increasing the neuron count in the first hidden layer from 128 to 256. We keep all other parameters the same, including the expanded data range and PER.

\begin{figure}[H]
    \centering
    \includegraphics[width=0.8\textwidth]{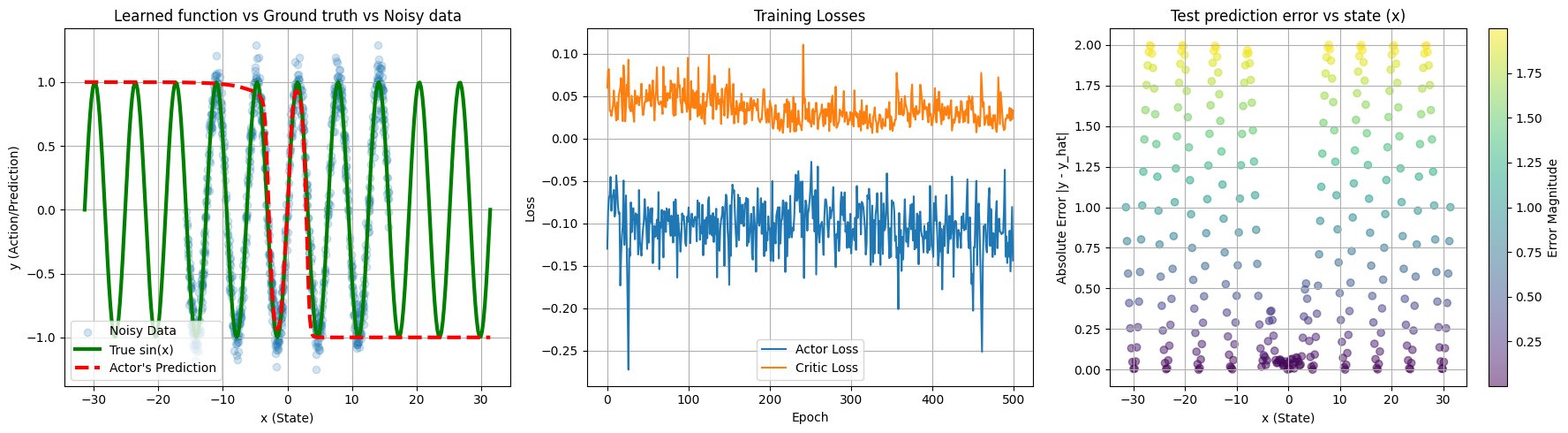}
    \caption{Stage 3 results. A deeper network begins to learn the pattern but fails to generalize to the edges of the distribution.}
    \label{fig:stage3}
\end{figure}

The results (Fig.\ref{fig:stage3}) show a plausible improvement. The model now attempts to learn the periodic function. It performs well in the center of the training regime but still struggles at the edges, failing to generalize to the outer cycles. While increasing network capacity was a necessary step, it was not sufficient on its own. The model still struggles to understand the underlying periodic structure of the data from the raw input $x$.

\subsection{The key ingredient: positional encoding}
The final and most significant improvement is to address the input state representation. The raw value of $x$ does not explicitly convey its periodic nature. To solve this, we introduce \textbf{positional encoding}, a feature engineering technique that transforms the scalar input $x$ into a higher-dimensional vector of sine and cosine functions with varying frequencies:
\begin{equation}
    PE(x)_i = 
    \begin{cases}
        \sin(2^{i/2} \cdot x) & \text{if } i \text{ is even} \\
        \cos(2^{(i-1)/2} \cdot x) & \text{if } i \text{ is odd}
    \end{cases}
\end{equation}
The dimensionality of the positional encoding vector was set to 16. This gives the model a rich, periodic representation of the input state. By feeding these features into the high-capacity network from Stage 3, the model can easily learn the mapping.

\begin{figure}[H]
    \centering
    \includegraphics[width=0.8\textwidth]{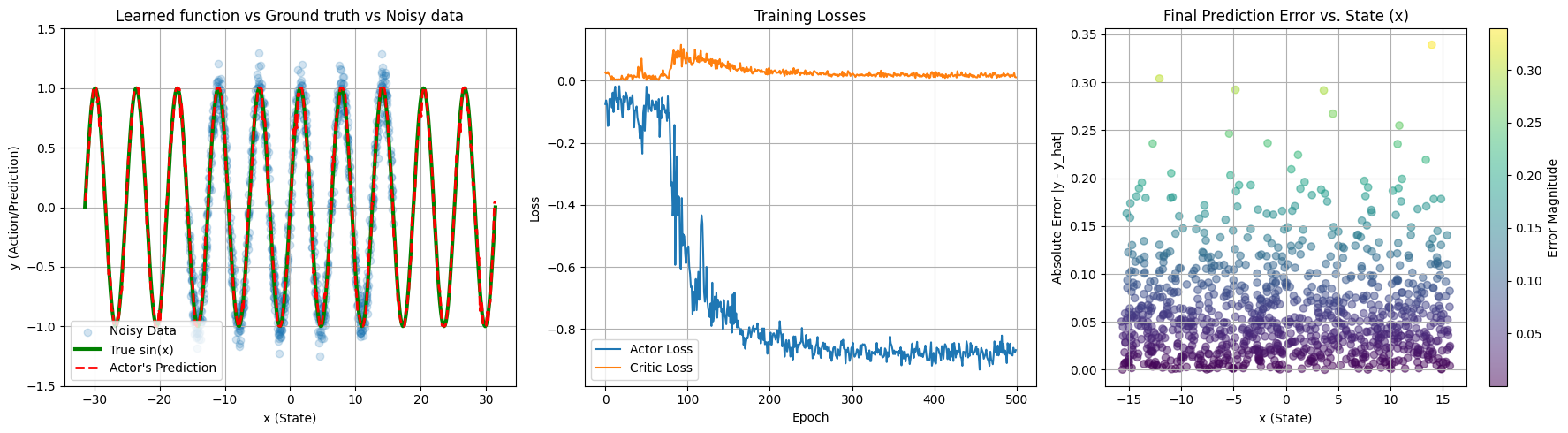}
    \caption{Stage 4 results. With positional encoding, the high-capacity network perfectly learns the function and generalizes correctly.}
    \label{fig:stage4}
\end{figure}

The results (Fig.\ref{fig:stage4}) are now perfect. The model accurately learns the sine function across all five periods of the training data and correctly extrapolates to the wider test range. The final error distribution is low and uniform, indicating that the combination of a powerful learning algorithm (Actor-Critic with PER), a high-capacity model, and informative features (positional encoding) has successfully solved the problem.

\section{Discussion}
This progressive example highlights several important lessons. First, framing regression as an RL problem is a viable and powerful approach. It grants us the flexibility to define custom, non-linear reward functions such as the Gaussian kernel, which can be more aligned with our true objectives than standard loss functions.
Second, the journey illustrates a standard workflow of tackling a periodic regression problem with RL. We started with a simple model, identified its failure modes, systematically addressed and fixed the issues. The failure in Stage 2 was due to a lack of \textbf{model capacity}, which we fixed by making the network larger. The failure in Stage 3 was due to a poor \textbf{input representation}, which we fixed with feature engineering (positional encoding) specific to the periodic data. The success of positional encoding is particularly noteworthy. It implies that the representation of features can be more critical than the learning algorithm itself, depending on the task. By supplying the model with features that were inherently aligned with the structure of the problem, the learning task became much simpler.

However, this approach is not without its drawbacks. RL algorithms are notably more complex to implement and tune than standard supervised learning methods. They are often more computational intensive, involve more hyper-parameters and can be less stable during training. For simple regression tasks where MSE is a sufficient objective, this RL framework would be overkill. Its true value lies in problems where the objective is complex, non-standard, or non-differentiable, and where the benefits of custom rewards and guided exploration outweigh the implementation overhead.

\section{Conclusion}
We empirically demonstrate that a regression task can be effectively re-framed and solved as a reinforcement learning problem. Using an Actor-Critic architecture, we trained an agent to perform function approximation by maximizing a custom reward signal. Our progressive case study on learning a sine wave illustrated a practical, iterative process for model development, moving from a simple, limited capacity model to a sophisticated and accurate one by systematically diagnosing failures and incorporating advanced techniques such as Prioritized Experience Replay, increased network capacity, and positional encoding. This work serves as a practical guide for researchers and practitioners interested in exploring the application of RL to problems traditionally handled by supervised regression, particularly in scenarios that demand flexible and highly customized objective functions. This RL as regression paradigm can be easily extended to RL for classification.
This 'RL as regressor' paradigm could be extended to classification, presenting an interesting direction for substituting supervised learning with RL.

\section*{Code Availability}
The code used in this work is available at: \url{https://github.com/YongchaoHuang/rl_regression}

\bibliographystyle{plain}
\bibliography{references}

\end{document}